\documentclass[11pt,letterpaper]{article}
\usepackage{emnlp2017}
\usepackage{times}
\usepackage{amsmath,amssymb}
\usepackage{latexsym}
\usepackage{multirow}
\usepackage{bm}
\usepackage{graphicx}
\usepackage{caption}
\usepackage{algorithm}
\usepackage[noend]{algpseudocode}

 \emnlpfinalcopy

\def\emnlppaperid{664}

\title{Later-Stage Minimum Bayes-Risk Decoding for Neural Machine Translation}

\author{
Raphael Shu \and Hideki Nakayama \\
\texttt{shu@nlab.ci.i.u-tokyo.ac.jp, nakayama@ci.i.u-tokyo.ac.jp} \\
\texttt{The University of Tokyo}
}

\date{\today}

\begin{document}

\maketitle

\begin{abstract}
For extended periods of time, sequence generation models rely on beam search as the decoding algorithm. However, the performance of beam search degrades when the model is over-confident about a suboptimal prediction. In this work, we enhance beam search by performing minimum Bayes-risk (MBR) decoding for some extra steps at a later stage. In our experiments, we found that the conventional MBR reranking is only effective with a large beam size. In contrast, later-stage MBR decoding is shown to work regardless of the choice of beam size, and outperform simple MBR reranking. Additionally, we found that the computation of Bayes risks can be much faster by calculating the discrepancies on GPU in batch mode.\end{abstract}

\section{Introduction}

Recently, neural-based sequence generation models have achieved state-of-the-art performance in machine translation \cite{wu2016google}. Neural machine translation (NMT) models normally utilize recurrent neural networks (RNNs) as decoders to generate output tokens sequentially. In the decoding phase, it is a common practice to use the beam search algorithm to find a candidate translation, which approximately maximizes the posterior probability.

Although beam search can find much better candidates compared to greedy decoding, sometimes it still produces inappropriate outputs. \citet{szegedy2016rethinking} has shown that a neural network can become too confident in a suboptimal prediction. In beam search, assigning high probability to a suboptimal prediction in one step may lead to a chain reaction that generates unnatural output sequences. 

To improve beam search, various approaches have been explored recently either by enhancing the scoring method \cite{li2016simple} or using reinforcement learning \cite{li2017learning,gu2017trainable}. In this work, we try to apply Minimum Bayes-Risk (MBR) decoding \cite{Kumar2004MinimumBD} to guide the decoding algorithm of NMT to find a better candidate. This approach exploits the similarity between candidate translations instead of predicting the quality of each single candidate.

In preliminary experiments, we found that simply reranking the results of NMT by their Bayes risks only works with a large beam size. However, we are facing two major difficulties when integrating the MBR reranking into beam search.


Firstly, we found that performing MBR reranking simultaneously with beam search does not work. MBR reranking shrinks the diversity of the candidate space, thus traps the decoding algorithm in a suboptimal search space. Moreover, as the probability of a partial translation cannot represent final confidence, the values of Bayes risks are inaccurate in the beginning.  


Secondly, as MBR reranking requires computing $N \times N$ discrepancy values to obtains the Bayes risks for $N$ candidates, CPU-based MBR reranking is excessively time-consuming.

In this paper, we propose to perform MBR decoding at a later stage to search for a ``refined'' candidate with low Bayes risk. To speed up MBR decoding, we designed two approaches to compute the Bayes risks on GPU, which are shown to be much faster than a standard implementation. The main contributions of this paper can be summarized as two folds:

\begin{enumerate}
  \item We found that MBR reranking only works with a large beam size. Conversely, performing MBR decoding at a later-stage is proved to be effective regardless of the choice of beam size, and outperform simple MBR reranking.
  \item We found that the computation of Bayes risks can be much faster by computing the discrepancy matrix on GPU in batch mode.
\end{enumerate}

\section{Related Work}

MBR decoding is widely applied in SMT \cite{Kumar2004MinimumBD,gonzalez2011minimum,duh2011generalized},  which is also found to improve the translation quality \cite{ehling2007minimum}. Recently, \citet{stahlberg2016neural} utilized the translation lattice of SMT to guide NMT decoding with the MBR decision rule, which is shown to be better than simply rescoring the N-best results of SMT. A drawback of this approach is that it requires a SMT system to be available and decode simultaneously with the NMT model.

Recently, some studies are proposed to enhance beam search with reinforcement learning. \citet{li2017learning} utilizes a simplified version of the actor-critic model to decode for arbitrary decoding objectives. The scoring function is modified to be an interpolation of the log probability and the output of the value function (or Q function). \citet{gu2017trainable} extends the noisy, parallel approximate decoding (NPAD) algorithm \cite{cho2016noisy} by adjusting the hidden states in recurrent networks with an agent, which is trained with the deterministic policy gradient algorithm \cite{Silver2014DeterministicPG}. These approaches predict the quality of each single candidate, whereas MBR reranking considers the relation between multiple candidates. Therefore, they can be combined with our model to further improve the quality of output sequences.


\section{Minimum Bayes-Risk Decoding}

MBR decoding is a technique to find a candidate with the least expected loss \cite{bickel1977mathematical}. Following previous work in SMT \cite{Kumar2004MinimumBD}, given an evidence space $E$, the Bayes risk of a candidate $y$ is computed by:
\begin{align}
    R(y) &= \sum_{y^\prime \in E} \Delta(y, y^\prime) p(y^\prime|x) \:. \label{eqn:risk}
\end{align}

The term $\Delta(y, y^\prime)$ gives the discrepancy between two candidates, which is normally computed by using $1 - \mathrm{BLEU}(y, y^\prime)$ in machine translation. In this paper, we use smoothed BLEU \cite{smoothed_bleu}, and the probability $p(y^\prime|x)$ is calculated by a {\it softmax} over the average log probabilities of all candidates in $E$ given by a NMT model. Intuitively, a candidate gets low Bayes risk if it is similar to the candidates in the evidence space. 

\section{Later-stage MBR Decoding}
\algnewcommand{\Inputs}[1]{%
  \State \textbf{Inputs:}
  \Statex \hspace*{\algorithmicindent}\parbox[t]{.8\linewidth}{\raggedright #1}
}
\algnewcommand{\Initialize}[1]{%
  \State \textbf{Initialize:}
  \Statex \hspace*{\algorithmicindent}\parbox[t]{.8\linewidth}{\raggedright #1}
}

In this section, we propose a simple decoding strategy, which searches for low-risk hypotheses after finishing beam search. The basic idea is to utilize the results of beam search as an evidence space to guide the later-stage MBR decoding.

As the hypotheses\footnote{Each hypothesis is a tuple of the average log probability, candidate tokens, and the lastly computed hidden state of the decoder LSTM.} that were discarded by beam search provide good starting points for finding low-risk hypotheses, we begin the later-stage decoding from a selection of discarded hypotheses rather than from scratch. To do this, in each step of beam search, we save $B$ discarded hypotheses that are outside the ``beam'' to a hypothesis list $H$, where $B$ is the number of beam size. After beam search, we select top $B$ finished hypotheses\footnote{A hypothesis is finished when a ``EOS'' token is reached.}, and save them to an evidence space $E$.

\begin{algorithm}[t]
\caption{Later-stage MBR decoding}\label{alg:latermbr}
\begin{algorithmic}[1]
\Inputs{$B$ = beam size, $T$ = time budget}
\Initialize{$H \gets$ discarded hypotheses \\ $E \gets B$ finished candidates }
\vskip 2pt
\For{$t \gets 1 \textrm{ to } T$}
    \State sort $H$ by Eq.\ \ref{eqn:new_score} in descending order
    \State $S \gets$ pop $B$ hyps. from $H$ with highest scores
    \State $S^\prime \gets$ decode $S$ to get $B \times B$ new hyps.
    \State push finished hyps. in $S^\prime$ to $E$
    \State push unfinished hyps. in $S^\prime$ back to $H$
\EndFor
\State perform MBR reranking for $E$
\State \textbf{output} $E[1]$ 
\end{algorithmic}
\end{algorithm}

After we collected the discarded hypotheses in $H$ and the evidences in $E$ from beam search, we perform the later-stage MBR decoding for an extra $T$ steps to recover low-risk hypotheses. In our experiments, $T$ is fixed to be the number of input words.

In each extra step, we sort\footnote{As computing $S(y)$ for all hypotheses in $H$ is time-consuming, in practice, we only rerank top $3 \times B$ hypotheses with the highest average log probabilities.} the hypothesis list $H$ with a score function, which will be described later in Section \ref{sec:score}. Then we pick $B$ hypotheses with the highest scores from $H$, and generate the next words for them, resulting in $B \times B$ new hypotheses. If a hypothesis is finished, we add it to the evidence space $E$. Otherwise, it is put back to the hypothesis list $H$. 

Finally, after performing the later-stage MBR decoding for $T$ steps, we select the candidate translation with the lowest risk in $E$ as the output. The complete algorithm is summarized in Alg.\ \ref{alg:latermbr}.


\subsection{Score Function for Hypothesis Selection}
\label{sec:score}

In the later-stage MRB decoding, we desire to guide the algorithm to find a hypothesis with low Bayes risk under the evidence space $E$. To achieve this, we select the hypotheses in $H$ to decode according to a score function computed for each hypothesis $y$ in each extra step, which is computed as follows:
\begin{align}
     S(y) = \frac{1}{|y|} \log p(y|x)  - \alpha R(y) - \beta L(y). \label{eqn:new_score}
 \end{align}
where the first part of the equation is the average log probability. The risk term $R(y)$ is computed by Eq.~\ref{eqn:risk}.

However, reranking with $R(y)$ alone will over-penalize short hypotheses as they are certainly dissimilar to the finished hypotheses in $E$. Therefore, we add a length penalty term $L(y)$ to encourage the selection of short hypotheses, which is proportional to the number of remaining steps:
\begin{align}
L(y) = (T-t) \cdot |y| .
\end{align}

In the last step of MBR decoding when $t=T$, as $L(y)$ will be zero, the hypotheses are selected only by their confidence scores and risks.

\subsection{Fast Computation of Bayes Risk}

Unfortunately, computing the Bayes risk for $N$ candidates requires evaluating the discrepancy function $\Delta(y, y^\prime)$ by $N \times N$ times. The computation is excessively time-consuming with a CPU-based implementation, whose bottleneck is to compute the following discrepancy matrix:
\begin{align}
    \begin{bmatrix}
    \Delta(y^1, y^1) & \dots  & \Delta(y^1, y^N) \\
    \vdots & \ddots & \vdots \\
    \Delta(y^N, y^1) & \dots  & \Delta(y^N, y^N)
\end{bmatrix}
\end{align}

To tackle this problem, we designed two approaches to compute the discrepancy matrix efficiently on GPU: (1) compute BLEU values in batches with a sophisticated GPU-based implementation (2) approximate the discrepancy values with a neural net. The advantage of the approximation approach is that the implementation is independent of the chosen criterion of discrepancy. In practice, we found that reranking with approximated discrepancies performs as good as a standard reranker.

For the GPU-based BLEU computation, the trick is to construct a count vector that contains the counts of all unique n-grams to compute matches. The implementational details are described in the supplementary material.

The neural-based approach approximates true discrepancy values with a simple LSTM, which can be computed in batches naturally:
\begin{align}
    \bm{h_t} &= \mathrm{LSTM_\theta}(E(y_t), \bm{h_{t-1}}), \\
    \bm{h^\prime_l} &= \mathrm{LSTM_\theta}(E(y^\prime_l), \bm{h^\prime_{l-1}}), \\
    \Delta(y_{1:t}, y^\prime_{1:l}) &= \bm{h_t}^\top \bm{h^\prime_l} + v_\theta ^\top (\bm{h_t} + \bm{h^\prime_{l}}) + b_\theta \: . \label{equation:risknet}
\end{align}
where $E(\cdot)$ is the one-hot embedding a token.




%

\subsection{Dynamically Adjusting Weights}
\label{sec:dynamic}



In this section, we turn our attention to the scoring function in Eq.\ \ref{eqn:new_score}. 
Similar to \citet{li2016simple}, we apply the REINFORCE algorithm \cite{williams1992simple} to learn the optimal weights ($\alpha$ and $\beta$) for each input. The difference is that we do not discretize the weights to make a finite action space. Instead, we directly apply REINFORCE algorithm to learn a Gaussian policy with continuous actions. The merit of this approach is that we do not need to find the effective range of each weight value beforehand.

We use the last state of the backward encoder LSTM in the NMT model to represent an input sequence, which is denoted by $\bm{s}$. The stochastic policy is defined as:
\begin{align}
    \bm{\mu} &=  f(\bm{s}; \bm{\theta_{\mu}}) \label{eqn:mean} \\
    \bm{\sigma} &=  \mathrm{softplus}(f(\bm{s}; \bm{\theta_{\sigma}})) \label{eqn:var} \\
     \pi(\bm{a}|\bm{s}; \bm{\theta}) &=  \prod_{i=1}^{|A|}\frac{1}{\sigma_i \sqrt{2 \pi}} \exp\big(-\frac{(a_i - \mu_i)^2}{2 \sigma_i^2}\big) \label{eqn:policy}
\end{align}
where $|A|$ is the number of actions, which equals 2 in our case. The function approximators $f(\cdot)$ are implemented with simple two-layer neural networks. In the training time, we sample actions from the distribution defined by the policy $\pi(\bm{a}|\bm{s}; \bm{\theta})$, whereas in the test time, the actions are computed deterministically with $f(\bm{s}; \bm{\theta_{\mu}})$.
%
%
%

\section{Experiments}

We evaluate our proposed decoding strategy on English-Japanese translation task, using ASPEC parallel corpus \cite{NAKAZAWA16.621}. The corpus contains 3M training pairs and 1812 sentence pairs in the test set. We tokenize the sentences with ``tokenizer.perl'' for English side, and Kytea \cite{kytea} for Japanese side. The sizes of vocabulary are cropped to 80k and 40k respectively. In our experiments, we trained a NMT model with a standard architecture \cite{bahdanau2014neural}, which has 1000 units for both the embeddings and LSTMs.

\subsection{Fast Bayes-Risk Computation}

\begin{figure}[ht]
  \centering
  \includegraphics[width=0.5\textwidth]{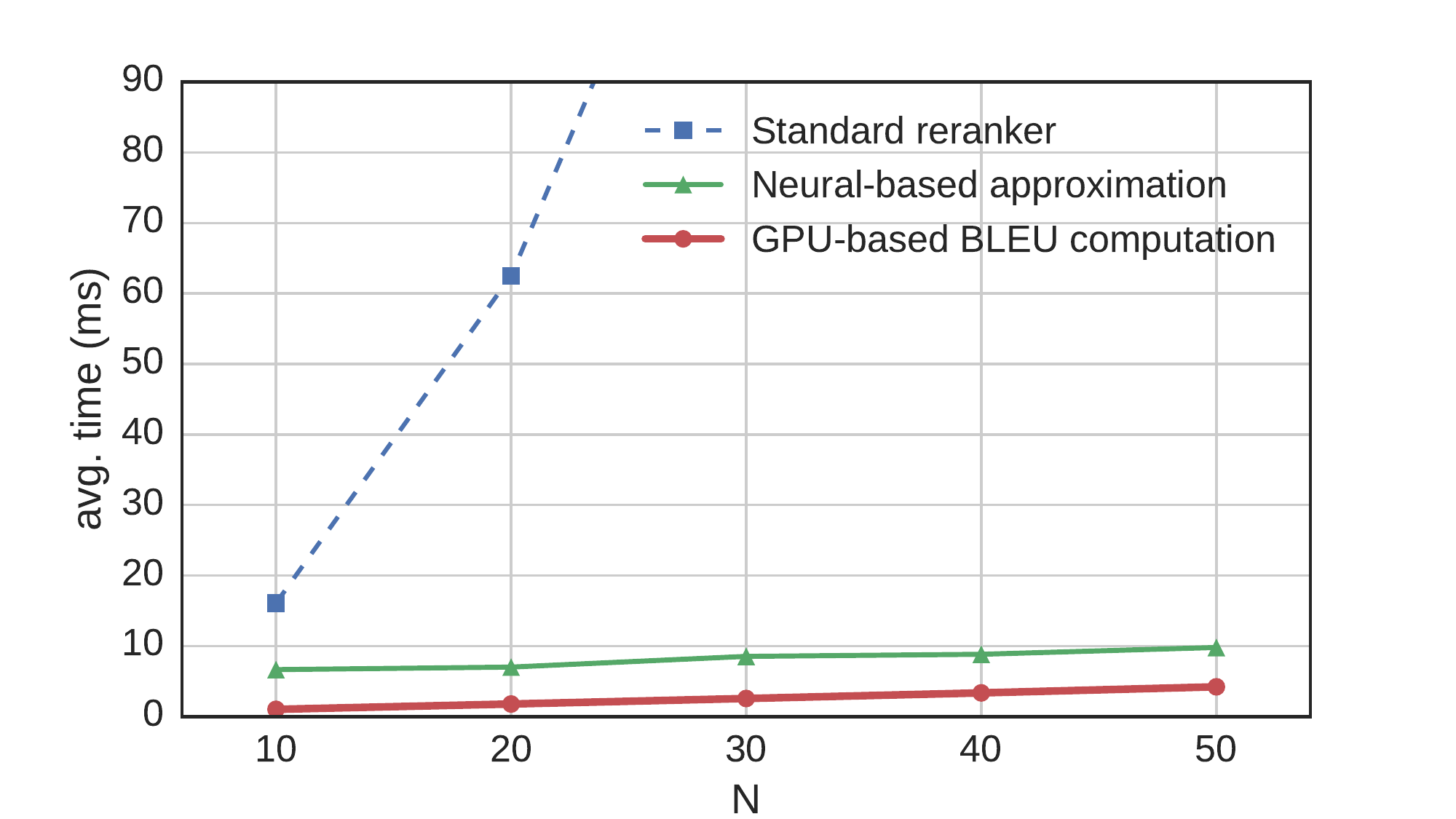}
  \caption{A comparison of reranking speed (time per sentence) with different approaches of computing Bayes risks. $N$ is the number of candidates. }
  \label{fig:rerank_spd}
\end{figure}

In this section, we compare the speed between the GPU-based Bayer-risk rerankers and a standard reranker. A comparison of reranking speed is shown in Fig. \ref{fig:rerank_spd}.



%
%
\captionsetup[figure]{skip=2pt}

For each input sentence, we rerank a list of $N$ candidate translations. The average reranking time per sentence is reported. The results show that both GPU-based approaches are much faster than a standard CPU-based reranker, thus capable of being integrated into beam search. Remarkably, the growing pattern of the average reranking time for GPU-based approaches is linear rather than exponential. 

In our experiment, reranking candidates with approximated discrepancy values is found to be as good as a normal reranker, shown in the middle of Table \ref{table:dynamic}. The training details are provided in supplementary material.

\subsection{Later-stage MBR decoding}

In Table \ref{table:dynamic}, we compare different decoding strategies in various settings of beam sizes. The BLEU scores are reported following a standard post-processing procedure\footnote{We produce the BLEU scores with Kytea tokenizer. The post-processing script can be found in \url{http://lotus.kuee.kyoto- u.ac.jp/WAT/} .}.

\begin{table}[ht]
  \begin{center}
  
    \begin{tabular}{r|c|c|c} \hline \hline
      \small{} & \multicolumn{3}{c}{\small{BLEU(\%)}} \\
      \cline{2-4}
      \small{} & \small{B=5} & \small{B=20} & \small{B=100} \\
      \hline
      \small{standard beam search (BS)} & 34.58 & 35.20 & 35.24\\
      \small{BS + MBR rerank}  & 34.56 & 35.35 &  35.65\\
      \small{BS + MBR rerank (approx)}  & 34.57 & 35.34 &  35.70\\
      \hline
      \small{BS + LaterMBR} &  \bf{34.96} & \bf{35.67} & \bf{35.87}\\  
      \hline \hline
    \end{tabular}
    \caption{Evaluation results of different decoding strategies. $B$ is the number of beam size.}
    \label{table:dynamic}
  \end{center}
\end{table}

We found that increasing the beam size to a large number does not consequently contribute to the evaluation scores. On the contrary, MBR reranking improves the scores when a large beam size is used, but less effective with a small beam size. Later-stage MBR decoding is shown to outperform the simple MBR reranking in all settings of beam size.

Additionally, we also found that the number of candidates in the evidence space largely affects the effectiveness of MBR reranking. In our experiments, the number of evidences is fixed to the same number of beam size. Using more evidences degrades the quality of selected candidates. 



\section{Conclusion and Future Work}

In this paper, we propose a simple decoding strategy to search a hypotheses with lowest Bayes risk at a later stage, which outperforms simple MBR reranking. We compute the Bayes risk on GPU to speed up step-wise MBR decoding.

Interestingly, we found the simple MBR reranking is especially effective with a large beam size. Without MBR reranking, further increasing the beam size does not result in significant gain. 

For future work, we intend to construct a better evidence space with an alternative neural network, in order to benefit the later-stage MBR decoding phase. 


\bibliography{mybib}
\bibliographystyle{emnlp_natbib}

\appendix

\section{Supplemental Materials}
\label{sec:supplemental}

\subsection{A Note on the Standard Implementation for Computing Bayes Risks}

As in the equation for computing $R(y)$, all the terms in the summation are positive numbers, we stop computing the risk of a candidate if the sum is already higher than the lowest risk value of computed candidates. Even with this early stopping technique, the standard MBR reranker still runs very slow as shown in Fig.~1 of the paper.

\subsection{GPU-based BLEU Computation}

For the particular discrepancy function based on BLEU, we found that a matrix of $N \times N$ BLEU values can be calculated efficiently on GPU. The trick is to build a $M$-dimensional count vector $\bm{c^i}$ for each candidate $y^i$, which contains the count of all $M$ uniq n-grams in the candidate space. Another vector $\bm{g}$ is used to indicate the rank of each n-gram.

For example, let a set of n-grams be \{``a'', ``in'', ``park'', ``ball'', ``in a'', ``a park''\}. Then for the sentence ``a park in a park'', $\bm{c^i}$ will be a 6-dimensional vector of $\{2, 1, 2, 0, 1, 2\}$, whereas $\bm{g}$ will be a vector of $\{1,1,1,1,2,2\}$.

The BLEU score of a candidate pair can be computed with:
\begin{align}
    &p^{i,j} = \frac{1}{4} \sum_{n=1}^4 \log \frac{\sum_{k=1}^M \mathrm{I}(g_k = n) \cdot \min(c_k^i, c_k^j) }{|y^i| + 1 - n} \label{eqn:bleu_p} \\
    &\mathrm{BLEU}(y^i, y^j) = \exp(\min(1 - \frac{|y^j|}{|y^i|}, 0) + p^{i,j})
\end{align}

In this way, a $N \times N$ BLEU matrix can be obtained in one shot with an input of $N \times M$ count matrix. In practice, we use smoothed BLEU, which simply adds 1 to both top and bottom parts of the fraction in Eq.\ \ref{eqn:bleu_p}.

\subsection{Training details of Neural-based Bayes-risk Approximation}

To learn a neural-based estimator that approximates the discrepancy matrix, we collect 100K candidate pairs by decoding the source sentences in the training data. For each input sentence, a new ID is assigned for each unique token in the candidate space to reduce the vocabulary size. For example, ``A cat eats'' and ``A dog eats'' will be converted to $\{0, 1, 3\}$ and $\{0, 2, 3\}$ respectively.

The model is trained with MSE loss function to predict $1-\mathrm{BLEU(y, y^\prime)}$ given a candidate pair. We use Adam optimizer \cite{kingma2014adam} and train the model for 50 epoch. The learning rate is fixed to $0.0001$. In practice, we scale the discrepancy scores by $0.1$. After 50 epochs, we obtain a MSE loss of $0.06$.

\subsection{A Note on Dynamic Weight Adjusting}

The model for predicting $\bm{\mu}$ and $\bm{\sigma}$ are both simple two-layer neural networks with a shared hidden layer of 100 units, followed by a tanh nonlinearity. 

The gradient for updating $\bm{\theta}$ by {\it gradient ascent} is given by the REINFORCE rule with a baseline:
\begin{align}
    \nabla&_{\bm{\theta}}J(\bm{\theta}) = \mathbb{E}_{\bm{a} \sim \pi} \Big[ (R_{s,a} - b_s) \nabla_{\bm{\theta}} \log \pi(\bm{a}|\bm{s}; \bm{\theta}) \Big] . \label{eqn:reinforce}
\end{align}

where $R_{s,a}$ is the reward for taking action $a$ given input $\bm{s}$, whereas $b_s$ is a baseline computed by another simple two-layer neural network. 
In practice, the computation of gradients used in the REINFORCE rule (Eq.~\ref{eqn:reinforce}) can be simplified as:
\begin{align}
    \nabla_{\bm{\theta}} \log \pi(\bm{a}|\bm{s}; \bm{\theta}) = \nabla_{\bm{\theta}} - \sum_{i=1}^{|A|} \frac{(a_i - \mu_i)^2}{2 \sigma_i^2} + \log \sigma_i \: . 
\end{align}

\end{document}